\documentclass[10pt,twocolumn,letterpaper]{article}

\usepackage[8]{cvpr} 

\definecolor{cvprblue}{rgb}{0.21,0.49,0.74}
\usepackage[pagebackref,breaklinks,colorlinks,allcolors=cvprblue]{hyperref}

\DeclareMathOperator*{\argmax}{arg\,max}

\usepackage{adjustbox}
\usepackage{multirow}
\usepackage{makecell}
\usepackage{pifont}
\usepackage{dsfont}

\def\paperID{10334} 
\def\confName{CVPR}
\def\confYear{2026}

\title{DPL: Decoupled Prototype Learning
for Enhancing Robustness of Vision–Language Transformers to Missing Modalities}

\author{Jueqing Lu\textsuperscript{1}, Yuanyuan Qi\textsuperscript{1}, Xiaohao Yang\textsuperscript{1},  
Shuaicheng Niu\textsuperscript{2},
Fucai Ke\textsuperscript{1},\\
Shujie Zhou\textsuperscript{1},
Wei Tan\textsuperscript{5},
Jionghao Lin\textsuperscript{3},
Wray Buntine\textsuperscript{4}
Hamid Rezatofighi\textsuperscript{1}, 
Lan Du\textsuperscript{1} \\
{\small \textsuperscript{1}Monash University}
{\small \textsuperscript{2}Nanyang Technological University}
{\small \textsuperscript{3}The University of Hong Kong}
{\small \textsuperscript{4}VinUni}
{\small \textsuperscript{5}Independent Researcher}
}

\begin{document}

\twocolumn[{
\begin{center}
\maketitle
\captionsetup{type=figure}
\includegraphics[width=\textwidth,]{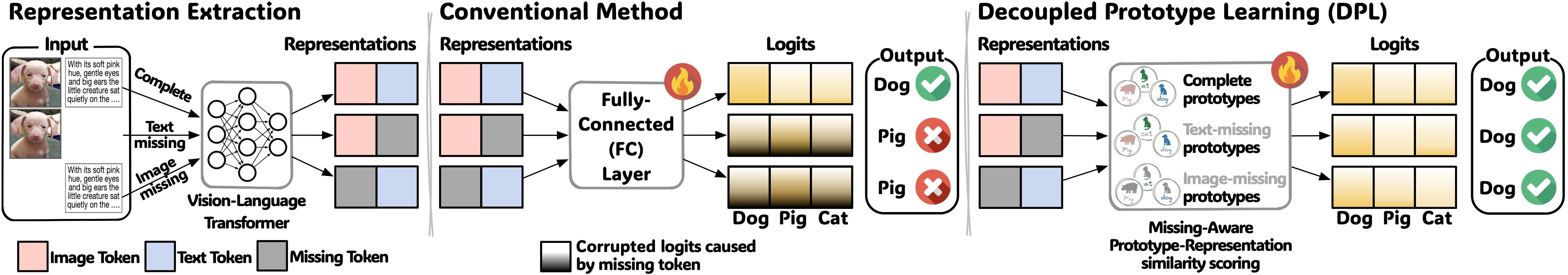}
\captionof{figure}{
Comparison between the conventional method (mid) and our proposed DPL framework (right) for multimodal classification under missing modalities. When some modalities are unavailable, the conventional method yield corrupted logits, whereas DPL uses missing-aware decoupling to mitigate the adverse impact of missing modalities.
}
\label{fig:dpl}
\end{center}
}]

\begin{abstract}The performance of Vision–Language Transformers drops sharply when an input modality (e.g., image) is missing, because the model is forced to make predictions using incomplete information. Existing missing-aware prompt methods help reduce this degradation, but they still rely on conventional prediction heads (e.g., a Fully-Connected layer) that compute class scores in the same way regardless of which modality is present or absent.
We introduce Decoupled Prototype Learning (DPL), a new prediction head architecture that explicitly adjusts its decision process to the observed input modalities. For each class, DPL selects a set of prototypes specific to the current missing-modality cases (image-missing, text-missing, or mixed-missing). Each prototype is then decomposed into image-specific and text-specific components, enabling the head to make decisions that depend on the information actually present. This adaptive design allows DPL to handle inputs with missing modalities more effectively while remaining fully compatible with existing prompt-based frameworks.
Extensive experiments on MM-IMDb, UPMC Food-101, and Hateful Memes demonstrate that DPL outperforms state-of-the-art approaches across all widely used multimodal image–text datasets and various missing cases.\end{abstract}    
\section{Introduction}
\label{sec:intro}

Driven by advances in Transformer architectures \cite{vaswani2017attention, devlin2018bert, dosovitskiy2020image, liu2021swin, kim2021vilt} and large-scale pretraining on paired corpora (\eg, image and text modalities) \cite{radford2021learning, kim2021vilt, li2022blip, li2023blip}, Vision-Language Transformers have achieved remarkable performance across diverse downstream tasks, such as image-text retrieval \cite{frome2013devise, lu2019vilbert, radford2021learning, li2022blip} and visual reasoning \cite{amizadeh2020neuro, thrush2022winoground, hsieh2023sugarcrepe, ke2024hydra, cai2025naver, ke2025dwim}.
Missing modality, a major obstacle to robustness and deployment of multimodal models, has sparked increasing research interest in recent years \cite{tran2017missing, zhou2021latent, ma2021smil, ma2022multimodal, lee2023cvpr, wang2023multi, kim2024missing, hu2024deep, liao2025benchmarking}, since it is inevitable in real-world multimodal systems due to practical limitations such as data collection constraints (\eg, withheld student grades in education) and privacy concerns (\eg, unavailable medical records in healthcare and privacy-driven data-sharing reluctance in personalized recommendations) \cite{prinsloo2023multimodal, khalid2023privacy, wang2018toward}.

The missing modality problem brings challenges not only by disrupting cross-modal interactions, but also by introducing noisy information into the model's decision-making process due to incomplete inputs (as in \cref{fig:dpl}), leading to unstable and degraded predictions.
This prompts us to ask: 
how can we fine-tune pretrained Vision-Language Transformers so that their prediction mechanisms remain adaptive under missing modalities?

Early efforts primarily adopted imputation or reconstruction strategies, attempting to recover missing modalities from the available ones \cite{tran2017missing, ma2021smil, wang2023multi, kim2024missing}. Such approaches suffered from two major limitations: 
1) they did not explicitly consider leveraging the knowledge in pretrained Transformers or how to efficiently adapt them for downstream tasks;
2) they relied heavily on multiple modalities, which restricts applicability when only a few modalities are available (\eg, vision-language tasks).
To address these limitations, more recent work has turned to prompt learning \cite{ma2022multimodal, lee2023cvpr, jang2024towards, dai2024muap, hu2024deep}, primarily focusing on vision-language Transformers due to their foundational role as the most prevalent and well-established multimodal architecture. Beyond addressing these limitations, prompts serve as contextual cues that mitigate the reliance of imputation-based methods on datasets with multiple complete modalities during training, enabling models to maintain robust performance even when downstream data naturally contains missing modalities.

However, current prompt-based methods only achieve partial adaptation to missing modalities. Despite using missing-aware prompts to guide representation extraction \cite{lee2023cvpr, jang2024towards, dai2024muap, hu2024deep}, their prediction heads remain uniform and insensitive to the specific missing case (\eg, image-missing case). This architectural inconsistency introduces a mismatch between the adaptive representations and the static prediction layer, preventing the model from fully exploiting missing-modality cues captured by the prompts. Recent studies \cite{wu2024comprehensive, wang2025topologically} have also highlighted that prediction-level adaptation to modality incompleteness remains underexplored. A truly adaptive approach should extend missing-awareness beyond representation adaptation to encompass the prediction strategy itself.

We propose \textbf{D}ecoupled \textbf{P}rototype \textbf{L}earning (DPL), a novel missing-aware prediction framework that learns tailored prediction heads for each modality configuration, effectively handling missing modalities during both training and testing. 
Instead of relying on a single set of class weights, DPL learns class-wise prototypes that are decoupled by missing cases and further decomposed by individual modalities. 
This design enhances model stability under missing modalities by adapting specialized prototypes to different missing cases. 
Importantly, DPL can be seamlessly integrated with existing prompt learning methods: while prompts improve representation adaptation, DPL enhances decision robustness through explicitly missing-aware and modality-specific prediction heads. 

We demonstrate that DPL consistently improves performance regardless of whether prompt fine-tuning is applied, highlighting its robustness and generalizability. 
Through extensive evaluation on three benchmark datasets (MM-IMDb, UPMC Food-101, and Hateful Memes), we show that integrating DPL into state-of-the-art prompt frameworks yields consistent improvements across various training and testing missingness scenarios.
Our contributions could be summarized as follows:
\begin{itemize}
    \item We introduce Decoupled Prototype Learning (DPL), an adaptive prediction head that selects class prototypes conditioned on the detected missing-modality case and performs classification through prototype–representation matching.

    \item We further enhance prototype learning by extending ArcFace loss with a missing-aware mechanism to learn prototypes that adapt to the available modalities, and introduce a Prototype Relational Contrastive Loss to regularize both inter-class and intra-class prototype relations, promoting semantic consistency and enhancing class separability.
    
    \item DPL works effectively as a standalone prediction head and integrates seamlessly with existing prompt-based methods, consistently outperforming state-of-the-art baselines across three multimodal benchmarks and multiple missing-modality scenarios.
\end{itemize}

\section{Related Work }

\textbf{Multimodal Learning with Missing Modalities.}
The performance degradation of multimodal models caused by missing modalities \cite{ma2022multimodal} has garnered increasing attention, as it raises critical concerns about the robustness of multimodal models.
Several methods have been proposed to enhance resilience of multimodal models.
MMIN \cite{zhao2021missing} mitigates missing modalities by predicting intermediate features from a shared multimodal representation learned from available modalities.
SMIL \cite{ma2021smil} employs a Bayesian meta-learning framework to estimate latent features for incomplete data, remaining effective even under high missing ratios.
\citet{ma2022multimodal} further improve Transformer robustness via multi-task learning with both complete and incomplete inputs, and automatically search for dataset-specific fusion strategies.
ShaSpec \cite{wang2023multi} aggregates information across samples through a shared prediction head, effectively compensating for missing inputs.
More recently, MAP \cite{lee2023cvpr} introduces missing aware prompts to handle missing modalities with minimal computational overhead, while MSP \cite{jang2024towards} and MuAP \cite{dai2024muap} extend MAP with modality-specific and multi-step prompt designs for better scalability.
DCP \cite{hu2024deep} further extend MAP by explicitly modeling the correlations among prompts and their inter-relationships with input features.

\begin{figure*}[t]
    \label{fig:framework_pic}
    \centering
    \includegraphics[width=0.9\linewidth]{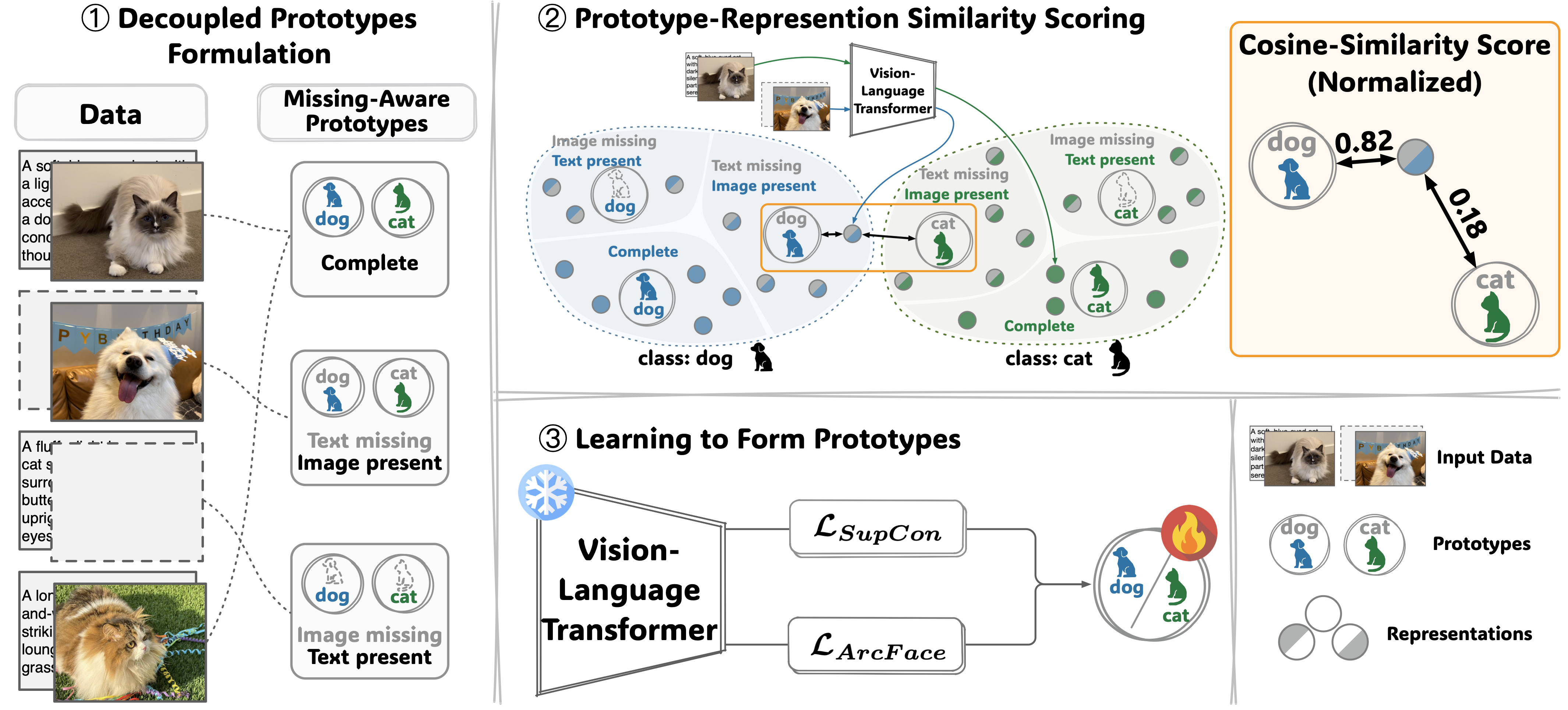}
    \caption{The workflow of DPL. 
    Class-wise prototypes are decoupled through a missing-aware mechanism. After feature extraction by the frozen pretrained encoder (with or without learnable prompts), the resulting features are matched with their corresponding prototypes to compute logits for predicting class labels. The prototypes are then updated based on the losses $\mathcal{L}_{\text{ArcFace}}$ and $\mathcal{L}_{\text{SupCon}}$.
    }
\end{figure*}

\noindent \textbf{Prompt Learning.} Prompt Learning, originally developed in NLP,
uses ``prompts'' to guide pretrained models for downstream tasks. 
Early methods \cite{petroni2019language, brown2020language} rely on manual prompts to improve model generalizability, 
while later approaches \cite{lester-etal-2021-power, li2021prefix} introduce learnable prompts that are optimized during training.
Afterwards, visual prompt tuning in Computer Vision has also made significant contributions \cite{jia2022vpt} as well as in test-time adaptation and continual learning \cite{lester-etal-2021-power, zhang2025dpcore,  xiao2025visual}.
Recently, prompt learning has been extended to multimodal tasks.
CoCoOp \cite{zhou2022conditional} extends CoOp's \cite{zhou2022learning} soft prompts with image-conditional prompts,
KgCoOp \cite{yao2023visual} aligns text embeddings with CLIP,
MaPLe \cite{khattak2023maple} integrates deep prompts into both image and text branches,
and DePT \cite{zhang2024dept} isolates base-specific knowledge during tuning.
These methods show the effectiveness of prompt learning in adapting vision-language models with minimal fine-tuning cost.
Building on this, MAP and its following work \cite{hu2024deep, lee2023cvpr, dai2024muap, jang2024towards} integrate prompt learning into multimodal models,
improving their robustness in missing-modality scenarios.

\noindent \textbf{Prototype Learning.}
Exemplar prototypes serve as references for model predictions and inferences, 
playing a core role in prototype learning.
This approach has demonstrated effectiveness in few-shot learning \cite{snell2017prototypical,dong2018few,li2021adaptive}, 
zero-shot learning \cite{xu2020attribute,jetley2015prototypical}, 
and has been successfully applied to face recognition \cite{deng2021variational,deng2019arcface,wang2018cosface,liu2019adaptiveface,huang2020curricularface}. 
Various loss functions have also been developed to improve prototypes learning \cite{schroff2015facenet,sohn2016improved,sun2020circle,deng2019arcface}.
Recently, conventional class-wise prototypes \cite{yang2018robust} have been extended to token-level prototypes 
for interpretable models \cite{chen2019looks,rymarczyk2022interpretable,nauta2021neural,ma2025interpretable} 
and sub-centroids for visual recognition \cite{wang2022visual}, 
leading to improved performance.
Our proposed DPL builds on the ArcFace Loss, explicitly dealing with missing modality cases and capturing
the correlations among prototypes.
This enhances the discriminability of learned prototypes 
and significantly improving the model's robustness to missing modalities, 
where a single shared classification head may not be sufficient.
Recent work \cite{li2025dpu} introduced prototype learning for Out-Of-Distribution (OOD) detection,
and concurrent to our work \cite{zhu2025dynamic} used prototypes for handling test-time domain shift.

\section{Decoupled Prototype Learning (DPL)}
\label{sec:method}

\subsection{Overall Workflow}
The overall workflow of DPL consists of three stages (\ding{192}–\ding{194}), as illustrated in \cref{fig:framework_pic}.
In stage \ding{192}, DPL constructs missing-aware, modality-specific prototypes for each class, including complete-modality, text-missing, and image-missing prototypes, as described in \cref{sec:de-prototypes}.
In stage \ding{193}, DPL conducts the forward pass with prototype–representation similarity scoring: given an input and its observed missing pattern, the corresponding prototype is dynamically selected, and class scores are computed via a similarity metric (\eg, cosine similarity), thereby aligning the decision boundary with the missing case conditioned representation, as discussed in \cref{sec:logits-compute}.
Finally, in stage \ding{194}, the pretrained Vision–Language Transformer backbone remains frozen, and the prototypes are optimized using the proposed loss functions introduced in \cref{sec:loss}.

\subsection{Problem Definition}
We build upon the commonly adopted formulation of missing-modality learning setting \cite{lee2023cvpr, jang2024towards, dai2024muap, hu2024deep}, and formalize a general setting where each sample may contain an arbitrary subset of available modalities.
For clarity, we illustrate the formulation with two modalities ($M = 2$), namely image ($I$) and text ($T$).

For a given multimodal dataset $D$,
its composition can include any pairwise combination of the following three sets,
as well as their complete union:
\{
$\mathcal{D}^{c}$,
$\mathcal{D}^{r_{I}}$,
$\mathcal{D}^{r_{T}}$
\},
where $\mathcal{D}^{c_{}}$ denotes the modality-complete subset,
and $\mathcal{D}^{r_{I}}$, $\mathcal{D}^{r_{T}}$ denote the image-missing subset and the text-missing subset, respectively.

For the multimodal classification task,
we define the these subsets as:
${\mathcal{D}}^{c}$ = \{$x^I$, $x^T$, $y$\},
${\mathcal{D}^{ r_{I} }}$ = \{$\tilde{x}^{I}$, $x^{T}$, $y$\},
and ${\mathcal{D}^{ r_{T} }}$ = \{$x^{I}$, $\tilde{x}^{T}$, $y$\}.
Here,
$x^{I}$ and $x^{T}$ are image and text inputs,
$\tilde{x}^{I}$ and $\tilde{x}^{T}$ are dummy inputs (\eg, empty pixels or strings) for missing cases,
and $y$ is the class label for either multi-class or multi-label classification.

\subsection{Decoupled Prototypes Formulation}
\label{sec:de-prototypes}
We employ decoupled prototypes for multi-modal classification in scenarios involving missing modalities.
Specifically, in a two-modality setup (\eg, image and text), 
we introduce three distinct prototypes:
$w_k^{c}$,
$w_k^{r_{I}}$
and 
$w_k^{r_{T}}$,
each tailored to instances belonging to $D^c$, $D^{r_{I}}$, or $D^{r_{T}}$, respectively, for the $k$-th class label.

We further introduce modality-tailoring mechanism to decompose each of these three prototypes for each class label.
The decomposition is formulated as:
{
    \small
    \begin{equation}
        w_{k}^{c} = [w_{I}^{c}, w_{T}^{c}];\
        w_{k}^{r_{I}} = [{w}_{k}^{r_{I}, I}, {w}_{k}^{r_{I}, T}];\
        w_{k}^{r_{T}} = [{w}_{k}^{r_{T}, I}, {w}_{k}^{r_{T}, T}]
    \label{eq:prototype}
    \end{equation}
}
For instance,
${w}_{k}^{r_{T},I}$ denotes the prototype of image modality (superscript $I$) of the $k$-th class (subscript $k$) when text modality is missing (superscript $r_{T}$).

It is worth noting that in prototype learning, prototypes will be normalized first before being used for logits computation. Without decomposition, normalization is applied to the entire prototype vector. In contrast, with decomposition, normalization is applied independently to each decomposed prototype component. This leads to different normalized prototypes and consequently different logits.

\subsection{Prototype-Representation Similarity Scoring}
\label{sec:logits-compute}

With regard to the multi-class classification task, we can classify an instance $x_i \in D$ to a specific label $\hat{y}_i$ based on $\hat{y}_i = \argmax_{k} ( \mathbf{z}_i )$,
where $\mathbf{z}_i$ is a $K$ dimensional vector that represents the logits of $i$-th data instance,
i.e., $\mathbf{z}_i = [z_{i, 1}, z_{i, 2}, \dots, z_{i, k}, \dots, z_{i, K}]$.
For the multi-label classification task, we apply a sigmoid function $\sigma(\cdot)$ to the logits vector,
and assign label $k$ to the instance when the corresponding probability surpasses a predefined threshold, \eg, 0.5.

Inspired by logits fusion across modalities \cite{li2023efficient} and zero-padding strategy for missing modalities \cite{hu2024deep}, we define the computation of the logits for class label $k$ as follows:
{
    \small
    \begin{align}
    \label{eq:logits}
        z_{i,k} &=  g([h_{i}^{I}, h_{i}^{T}], w_k^{\cdot}) \nonumber \\
        &= \begin{cases}
        \big( (\hat{h}_{i}^{I})^{\mathsf{T}} \hat{w}_{k}^{c, I} + (\hat{h}_{i}^{T})^{\mathsf{T}} \hat{w}_{k}^{c, T} \big) / 2  & \text{if } x_i \in D^c \\
        (\hat{h}_{i}^{T})^{\mathsf{T}} \hat{w}_{k}^{r_{I}, T} & \text{if } x_i \in D^{r_{I}} \\
        (\hat{h}_{i}^{I})^{\mathsf{T}} \hat{w}_{k}^{r_{T}, I} & \text{if } x_i \in D^{r_{T}} \\
        \end{cases}
    \end{align}
}
where $g$ denotes the logits computation based on prototype-representation similarity (\eg, cosine similarity).
$\hat{h}_{i}^{I}$ and $\hat{h}_{i}^{T}$ are L2-normalized representation of the image modality $h_{i}^{I}$ and the text modality $h_{i}^{T}$, respectively. Similarly, $\hat{w}_{k}^{\cdot}$ denotes the L2-normalized prototype of $w_{k}^{\cdot}$.

It is worth noting that the representation of the missing modality is replaced by a zero-padded vector, same as in \cite{hu2024deep}, ensuring that only the available modality contributes to the logits. For example, when the image modality is missing ($r_I$), the normalized text prototype $\hat{w}_{k}^{r_{I}, T}$ is used with the normalized text representation $\hat{h}_{i}^{T}$. This computation maintains consistency with the magnitude of logits from complete-modality samples by averaging over both modalities when available.

\subsection{Learning to Form Decoupled Prototypes}
\label{sec:loss}

\noindent \textbf{Overall Training Objective.} We summarize the overall training objective of DPL as the following loss term:
{
    \small
    \begin{equation}
    \mathcal{L}_{\text{DPL}} = \mathcal{L}_{\text{ArcFace}} + \lambda \mathcal{L}_{\text{PRC}}
    \end{equation}
}
where $\lambda$ is the balancing coefficient.

\noindent \textbf{Adaptive Prototype Optimization $\mathcal{L}_{\text{ArcFace}}$.}
We optimize the decoupled prototypes using a margin-based classification objective inspired by ArcFace \cite{deng2019arcface}, with adaptive parameters for learning under missing modalities.
The additive angular margin encourages effective separation between classes, but the original formulation assumes uniform data quality, which does not hold when modality availability varies.
We therefore adapt this framework to our decoupled prototype design, enabling optimization that is conditioned on the observed modalities without altering the pretrained backbone.
To further reflect the confidence of different modality configurations, we assign adaptive scale and margin parameters $(s,m)$ to each prototype type: complete, image missing, and text missing. Formally, each subset of prototypes ${w^{c}, w^{r_I}, w^{r_T}}$ is trained with its own scale and margin $(s^{c},m^{c})$, $(s^{r_I},m^{r_I})$, and $(s^{r_T},m^{r_T})$.
This design balances decision boundaries across diverse missing cases, prevents the dominance of complete modality data, and improves robustness under missing inputs.

\noindent \textbf{Prototype Relational Contrastive Loss $\mathcal{L}_{\text{PRC}}$. } 
The decoupled prototypes of each class are trained on different subsets (\eg, complete or missing-modality data) and may drift apart if optimized independently.
To maintain semantic coherence, we introduce the Prototype Relational Contrastive Loss that aligns prototypes of the same class while separating those of different classes.
The loss is formulated as
{
    \small
    \begin{equation}
    \mathcal{L}_{\text{PRC}} = - \sum_{k=1}^{K} \sum_{u,v} \mathds{1}_{u \neq v} \log \frac{\exp(\hat{w}_k^{u} \cdot \hat{w}_k^{v})}{\sum}
    \end{equation}
}
where $\sum$ = $\sum_{l=1}^{K} \sum_{o} \mathbb{I}_{(l = k \land u \neq o) \lor (l \neq k)} \exp(\hat{w}_k^{u} \cdot \hat{w}_l^{o})$ is the normalization factor.
The superscripts $u$, $v$, $o$ each independently index over all combinations of missing pattern and modality, \eg, $(r_{T},I)$ for the prototype of image modality ($I$) for the text-missing case ($r_{T}$).
\section{Experiments}
\subsection{Experiment Setup}

\textbf{Datasets.}
Following \cite{ma2022multimodal, lee2023cvpr, hu2024deep},
we evaluated our DPL on three benchmark datasets:
\begin{itemize}
    \item \textbf{MM-IMdb} is a multimodal dataset for multi-label movie genre classification, where each movie is associated with multiple genres. It is among the largest publicly available datasets for this task, containing both image and text modalities.
    \item \textbf{UPMC Food-101} is an extension of the ETHZ Food-101 dataset, augmented with textual descriptions collected from Google Image Search. It retains 101 food categories but contains noisy image–text pairs due to its web-sourced nature.
    \item \textbf{Hateful Memes} is a challenging multimodal benchmark developed by Facebook AI for hate speech detection in memes. It is specifically designed to prevent models from relying on a single modality (either text or image).
\end{itemize}

\begin{table*}[t!]
    \caption{
    Test results on different datasets with various missing rates and missing cases of different methods by using CLIP as backbone. 
    FC means fully-connected layer.
    `source' indicates the original implementation in the related work. The backbone will keep frozen during training, while prompts (if applicable), the FC layer and decoupled prototypes will be fine-tuned.
    Best results are highlighted in \textbf{bold}.
    }
    \centering
    \small
    \adjustbox{max width=0.95\textwidth}{
    \begin{tabular}{c|c|cc|cc|cc|cc|cc}
        \hline 
          \multirow{3}{*}{ Datasets } & \multirow{3}{*}{\makecell[c]{  Missing \\ rate $\eta$}} & \multicolumn{2}{c|}{ Train/Test} & \multicolumn{2}{c|}{w/o Prompt Tuning} & \multicolumn{2}{c|}{w/ MaPLe (CVPR'23)} & \multicolumn{2}{c|}{w/ MAP (CVPR'23)}  & \multicolumn{2}{c}{w/ DCP (NeurIPS'24)} \\
           & & \multirow{2}{*}{Image} & \multirow{2}{*}{Text}   & w/ FC & w/ DPL & w/ FC & w/ DPL & w/ FC & w/ DPL & w/ FC & w/ DPL \\
           & &  &    & (source) & (ours) & (source) & (ours) & (source) & (ours)  & (source) & (ours) \\
          \hline 
          \multirow{9}{*}{\makecell[c]{ 
          MM-IMDb \\(F1-Macro)}} & \multirow{3}{*}{$50 \%$}  & $100 \%$ & $50 \%$  & 54.38 & \textbf{56.17} & 54.31 & \textbf{56.36} & 53.32 & \textbf{56.03} & 53.62 & \textbf{56.94}\\
                                  &                          & $50 \%$ & $100 \%$  & 55.39 & \textbf{58.33} & 56.90 & \textbf{59.01} & 56.97 & \textbf{58.91} & 56.50 & \textbf{58.70}\\
                                  &                          & $75 \%$ & $75 \%$   & 55.08 & \textbf{56.87} & 54.47 & \textbf{57.00} & 54.55 & \textbf{57.00} & 55.21 & \textbf{57.24}\\
          \cline{2-12}
                                  & \multirow{3}{*}{$70 \%$} &  $100 \%$ & $30 \%$ & 52.28 & \textbf{54.29} &  50.87 & \textbf{54.50} & 50.90 & \textbf{53.43} &  51.35 & \textbf{54.07}\\
                                  &                          & $30 \%$ & $100 \%$  & 54.12 & \textbf{57.30} &  55.99 & \textbf{58.56} & 55.10 & \textbf{58.17} &  55.87 & \textbf{57.69}\\
                                  &                          & $65 \%$ & $65 \%$   & 52.02 & \textbf{54.91} &  51.66 & \textbf{55.44} & 51.75 & \textbf{54.85} &  51.46 & \textbf{55.48}\\
          \cline{2-12}
                                  & \multirow{3}{*}{$90 \%$} & $100 \%$ & $10 \%$  & 48.84 & \textbf{52.90} &  49.17 & \textbf{52.85} & 49.20 & \textbf{52.55} &  49.69 & \textbf{53.06}\\
                                  &                          & $10 \%$ & $100 \%$  & 52.60 & \textbf{55.93} &  54.57 & \textbf{57.50} & 54.66 & \textbf{58.02} &  55.07 & \textbf{57.35}\\
                                  &                          & $55 \%$ & $55 \%$   & 50.24 & \textbf{53.98} &  51.11 & \textbf{54.66} & 49.98 & \textbf{53.90} &  51.00 & \textbf{54.73}\\
         \hline 
         \multirow{9}{*}{\makecell[c]{ 
          UPMC \\ Food-101 \\(Accuracy)}} & \multirow{3}{*}{$50 \%$}  & $100 \%$ & $50 \%$ & 82.21 & \textbf{84.99} & 82.45 & \textbf{85.08} & 82.10 & \textbf{85.37} &  82.18 & \textbf{85.14}\\
                                  &                          & $50 \%$ & $100 \%$          & 88.51 & \textbf{89.41} & 89.18 & \textbf{89.92} & 89.15 & \textbf{89.89} &  89.36 & \textbf{89.83}\\
                                  &                          & $75 \%$ & $75 \%$           & 84.76 & \textbf{86.78} & 84.94 & \textbf{87.31} & 85.13 & \textbf{87.08} &  85.20 & \textbf{87.15}\\
          \cline{2-12}
                                  & \multirow{3}{*}{$70 \%$} &  $100 \%$ & $30 \%$         & 79.57 & \textbf{82.04} & 79.58 & \textbf{82.21} & 79.66 & \textbf{82.08} &  79.53 & \textbf{81.88}\\
                                  &                          & $30 \%$ & $100 \%$          & 86.98 & \textbf{87.59} & 87.80 & \textbf{88.11} & 87.51 & \textbf{88.23} &  87.56 & \textbf{88.06}\\
                                  &                          & $65 \%$ & $65 \%$           & 82.22 & \textbf{84.26} &  82.16 & \textbf{84.25} & 82.29 & \textbf{84.43} &  82.13 & \textbf{84.24}\\
          \cline{2-12}
                                   & \multirow{3}{*}{$90 \%$} & $100 \%$ & $10 \%$         & 77.10 & \textbf{78.90} &  76.51 & \textbf{79.05} & 77.21 & \textbf{78.93} &  76.37 & \textbf{78.90}\\
                                  &                          & $10 \%$ & $100 \%$          & 85.98 & \textbf{86.12} &  86.23 & \textbf{86.39} & 86.39 & \textbf{86.49} &  86.26 & \textbf{86.68}\\
                                  &                          & $55 \%$ & $55 \%$           & 79.44 & \textbf{81.99} &  79.60 & \textbf{81.81} & 79.54 & \textbf{81.98} &  79.16 & \textbf{81.75}\\
        \hline 
          \multirow{9}{*}{\makecell[c]{  
          Hateful-\\Memes \\(AUROC)}} & \multirow{3}{*}{$50 \%$}     & $100 \%$ & $50 \%$  & 69.17 & \textbf{71.61} &  69.74 & \textbf{70.90} & 66.72 & \textbf{69.93} &  69.84 & \textbf{70.87}\\
                                  &                          & $50 \%$ & $100 \%$          & 66.61 & \textbf{67.46} &  64.83 & \textbf{66.87} & 63.89 & \textbf{64.53} &  65.51 & \textbf{66.18}\\
                                  &                          & $75 \%$ & $75 \%$           & 69.00 & \textbf{69.31} &  67.37 & \textbf{68.71} & 67.16 & \textbf{68.08} &  66.85 & \textbf{69.18}\\
          \cline{2-12}
                                  & \multirow{3}{*}{$70 \%$} &  $100 \%$ & $30 \%$         & 68.33 & \textbf{70.57} &  69.42 & \textbf{70.27} & 68.76 & \textbf{70.35} &  69.45 & \textbf{70.24}\\
                                  &                          & $30 \%$ & $100 \%$          & 65.42 & \textbf{65.62} &  63.02 & \textbf{64.42} & 60.87 & \textbf{63.69} &  62.62 & \textbf{65.51}\\
                                  &                          & $65 \%$ & $65 \%$           & 67.34 & \textbf{67.68} &  65.59 & \textbf{66.87} & 64.99 & \textbf{66.83} &  64.96 & \textbf{67.68}\\
          \cline{2-12}
                                  & \multirow{3}{*}{$90 \%$} & $100 \%$ & $10 \%$          & 66.61 & \textbf{70.76} &  68.84 & \textbf{70.42} & 67.57 & \textbf{69.85} &  68.19 & \textbf{70.65}\\
                                  &                          & $10 \%$ & $100 \%$          & 63.48 & \textbf{63.60} &  60.78 & \textbf{61.87} & 62.88 & \textbf{63.02} &  62.46 & \textbf{62.74}\\
                                  &                          & $55 \%$ & $55 \%$           & 64.65 & \textbf{66.40} &  62.94 & \textbf{65.00} & 62.07 & \textbf{64.96} &  64.28 & \textbf{65.45}\\
        \hline
         
    \end{tabular}
    }
    
    \label{table:main-results}
\end{table*}

\noindent \textbf{Evaluation Metrics.}
We adopt the same evaluation metrics as in \cite{ma2022multimodal, lee2023cvpr, hu2024deep}.
For MM-IMDb, we report Macro-F1 for multi-label classification.
For UPMC Food-101, we use Top-1 accuracy.
For the binary classification task on Hateful Memes, we report the Area Under the ROC Curve (AUROC).

\noindent \textbf{Setting of Missing Modality.}
Following prior works \cite{lee2023cvpr, hu2024deep}, we simulate missing modalities during both training and testing using a missing rate $\eta$, which controls the proportion of incomplete samples.
In the image-missing scenario ($D$ = $D^{C} \cup D^{R_I}$),
$\eta$ indicates that $\eta\%$ of samples contain text only, while the remaining $1-\eta$\% are complete.
Similarly, 
In the text-missing scenario ($D$ = $D^{C} \cup D^{R_T}$), $\eta$ represents the proportion of image-only instances, and 1-$\eta$\% are complete.
For the mixed-missing scenario ($D$ = $D^{C} \cup D^{R_I} \cup D^{R_T}$), the dataset includes $\frac{\eta}{2}$\% text-only samples, $\frac{\eta}{2}$\% image-only samples, and 1-$\eta$\% complete samples.
To assess robustness, we varied $\eta$ among 50\%, 70\% and 90\%.

\noindent \textbf{Comparison Methods.}
To assess the effectiveness of DPL, we first evaluate it against a conventional Fully-Connected (FC) layer.
As DPL can be seamlessly integrated with existing prompt tuning approaches, we further combine it with several state-of-the-art (SOTA) methods, including MaPLe, MAP, and DCP.
Following DCP, we also reimplement MAP with consistent backbones for fair comparison, and evaluate DPL with both the ViLT backbone used in MAP and the CLIP backbone used in DCP.
Additionally, we compare DPL with DePT, a SOTA method that decouples task-specific knowledge across feature channels to improve generalization in prompt tuning.

\noindent \textbf{Implementation Details.}
All baselines were primarily reimplemented using the official DCP code with a CLIP backbone.
For CLIP, we use the ViT-B/16 visual encoder with input images resized to $224 \times 224$.
The text encoder is inherited from the pretrained CLIP model with a maximum token length of 77.
All pretrained encoder parameters are frozen, except for learnable prompts (if applicable) and the proposed prototypes.
The prompt length is set to 36, consistent with DCP.
We apply the same configurations to other compared methods, including MaPLe and MAP.
For optimization, we follow \cite{hu2024deep, lee2023cvpr} and use the AdamW optimizer \cite{loshchilov2017decoupled} with an initial learning rate of $1\times10^{-2}$ and weight decay of $2\times10^{-2}$.
The learning rate is warmed up for the first 10\% of training steps and then decayed linearly to zero.
Following DCP, we use a batch size of 4 on a single RTX 3090 GPU.
Missing modalities are replaced by zero-filled tensors.
Additionally, we evaluate DPL using the official MAP implementation, which employs a ViLT backbone and 16-token prompts, on the Hateful Memes dataset (see \cref{table:map-official}).
This evaluation ensures fairness given backbone differences and the smaller test split in prior work.

\subsection{Effectiveness Analysis}

\textbf{Consistent Enhancement.} 
The proposed DPL consistently improves existing baselines by replacing the conventional prediction head, i.e., the Fully-Connected (FC) layer, regardless of whether soft prompts are used during fine-tuning.
As summarized in \cref{table:main-results}, we directly compare the FC and DPL variants across multiple baselines and highlight the superior results in \textbf{bold}.
DPL achieves consistent performance gains over all baselines. 
In particular, when prompt tuning is disabled (i.e., w/o Prompt Tuning), DPL provides a clear improvement since only the prediction head is updated while the pretrained backbone remains frozen.
The improvement is observed across all three benchmarks and under various missing rates (\eg, 50\%, 70\%, and 90\%) and missing cases (single-modality or mixed missing).
Even under severe modality loss (\eg, 90\%), DPL effectively mitigates performance degradation, improving the baseline by up to 1.8\% when both image and text inputs are partially missing.

\noindent \textbf{Unleash the Potential of Prompt-Guided Representations.} To further validate the effectiveness of DPL, we integrate it with existing prompt-based methods by replacing FC layer.
As shown in \cref{table:main-results}, DPL consistently enhances the performance of prompt-tuning baselines.
For instance, in the 50\% text-missing case (the first row in the table), DPL improves MaPLe, MAP, and DCP from 54.31\% $\rightarrow$ 56.36\% ($\Delta$=+2.05\%), 53.32\% $\rightarrow$ 56.03\% ($\Delta$=+2.71\%) and 53.62\% $\rightarrow$ 56.94\% ($\Delta$=+3.32\%), respectively.
These consistent improvements demonstrate that DPL effectively unlocks the potential of prompt-guided representations, complementing existing prompt-tuning methods by providing a more adaptive and discriminative prediction space.

\begin{table}[t]
    \caption{Test AUROC on Hateful-Memes with different backbones (ViLT and CLIP) and prediction heads (prompt lengths inherited from MAP and DCP) upon MAP. \textbf{\textcolor{blue}{bold}}/\textbf{bold} for the best.}
    \centering
    \adjustbox{max width=\linewidth}{
    \begin{tabular}{cc|cccc}
        \hline 
        \multicolumn{2}{c|}{ Train/Test}  & \multicolumn{4}{c}{MAP (CVPR'23)}  \\
         \multicolumn{2}{c|}{ $\eta$=70\% } & \textcolor{blue}{16 prompts}  & \textcolor{blue}{16 prompts} & 36 prompts  & 36 prompts  \\
        {Image} & {Text}   & \textcolor{blue}{ViLT(FC)} & \textcolor{blue}{ViLT(DPL)} & CLIP(FC) & CLIP(DPL)  \\
        \hline 
        $100\%$ & $30  \%$ & \textcolor{blue}{62.24} & \textbf{\textcolor{blue}{63.29}} & 68.76 & \textbf{70.35} \\
        $30 \%$ & $100 \%$ & \textcolor{blue}{54.02} & \textbf{\textcolor{blue}{64.53}} & 60.87 & \textbf{63.69} \\
        $65 \%$ & $65  \%$ & \textcolor{blue}{59.48} & \textbf{\textcolor{blue}{62.08}} & 64.99 & \textbf{66.83} \\
        \hline
    \end{tabular}
    }
    \label{table:map-official}
\end{table}

\noindent \textbf{Adaptivity to Different Backbones.} Although CLIP is widely adopted, we recognize that other backbone architectures may exhibit distinct characteristics.
To provide a comprehensive evaluation of the proposed DPL, we further test its effectiveness by replacing the CLIP backbone with ViLT.
Unlike CLIP’s dual-stream design, ViLT employs a single-branch architecture that directly fuses visual and textual tokens.
Since ViLT serves as the official backbone of MAP, this setup allows us to assess the complementarity of DPL within the MAP framework.
As shown in \cref{table:map-official}, our DPL demonstrates strong adaptability to different backbones, consistently improving performance across various missing-modality patterns.
We set the missing rate to $\eta\%$=70, following the standard configuration in the MAP paper, and use the same prompt lengths as defined in the MAP and DCP implementations.

\begin{table}[t]
    \caption{Comparison between FC, DePT and DPL upon two main baselines (MAP and DCP) on UPMC Food-101 dataset. The best and the second best results are highlighted in \textbf{bold} and \underline{underline}.}
    \centering
    \adjustbox{max width=0.48\textwidth}{
    \begin{tabular}{cc|ccc|ccc}
        \hline 
        \multicolumn{2}{c|}{ Train/Test} & \multicolumn{3}{c|}{MAP (CVPR'23)} & \multicolumn{3}{c}{DCP (NeurIPS'24)} \\
        \multicolumn{2}{c|}{ $\eta$=\textcolor[HTML]{007acc}{50}/\textcolor[HTML]{0a8842}{70}/\textcolor[HTML]{B66F1B}{90}\% } & Original & DePT & DPL & Original & DePT & DPL \\
        Image &  Text  & Head & (CVPR'24) & (ours) & Head & (CVPR'24) & (ours) \\
        \hline 
        \textcolor[HTML]{007acc}{100}\% & \textcolor[HTML]{007acc}{50}\%  & 82.10 & \underline{82.15} & \textbf{85.37} & 82.18 &\underline{82.37}& \textbf{85.14} \\
        \textcolor[HTML]{007acc}{50}\%  & \textcolor[HTML]{007acc}{100}\% & \underline{89.15} & 87.05 & \textbf{89.89} & \underline{89.36} &87.35& \textbf{89.83} \\
        \textcolor[HTML]{007acc}{75}\% & \textcolor[HTML]{007acc}{75}\% & \underline{85.13} & 84.02 & \textbf{87.08} & \underline{85.20} &84.69& \textbf{87.15} \\
        \cline{1-8}
        \textcolor[HTML]{0a8842}{100}\% & \textcolor[HTML]{0a8842}{30}\%  & \underline{79.66} & 79.63 & \textbf{82.08} & \underline{79.53} &79.39& \textbf{81.88} \\
        \textcolor[HTML]{0a8842}{30}\% & \textcolor[HTML]{0a8842}{100}\% & \underline{87.51} & 86.27 & \textbf{88.23} & \underline{87.56} &85.83& \textbf{88.06} \\
        \textcolor[HTML]{0a8842}{65}\% & \textcolor[HTML]{0a8842}{65}\% & \underline{82.29} & 82.07 & \textbf{84.43} & 82.13 &\underline{82.16}& \textbf{84.24} \\
        \cline{1-8}
        \textcolor[HTML]{B66F1B}{100}\% & \textcolor[HTML]{B66F1B}{10}\%  & \underline{77.21} & 76.98 & \textbf{78.93} & 76.37 &\underline{76.97}& \textbf{78.90} \\
        \textcolor[HTML]{B66F1B}{10}\% & \textcolor[HTML]{B66F1B}{100}\%  & \underline{86.39} & 85.30 & \textbf{86.49} & \underline{86.26} &85.06& \textbf{86.68} \\
        \textcolor[HTML]{B66F1B}{55}\% & \textcolor[HTML]{B66F1B}{55}\%  & 79.54 & \underline{79.97} & \textbf{81.98} & 79.16 &\underline{79.50}& \textbf{81.75} \\
        \hline
         
    \end{tabular}
    }
    \label{table:dept-results}
\end{table}
\begin{figure*}[t]
    \label{fig:fine-grained-missing}
    \centering
    \includegraphics[width=0.9\linewidth]{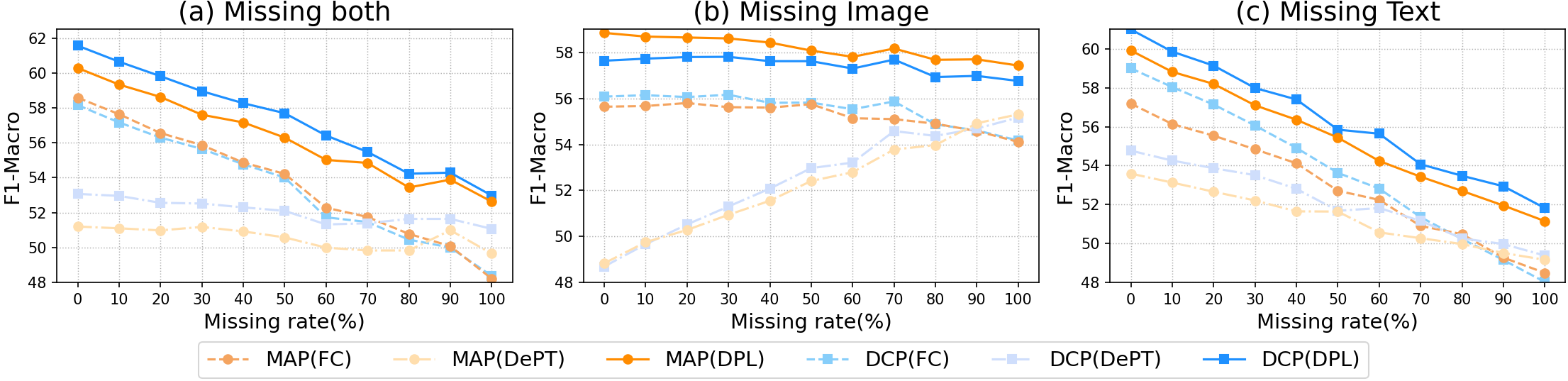}
    \caption{F1-Macro comparison of MAP and DCP frameworks with different prediction heads (FC, DePT, DPL) on MM-IMDb under varying missing rates. The proposed DPL consistently outperforms others across all missing-modality scenarios: (a) both missing, (b) image missing, and (c) text missing.}
\end{figure*}


\noindent \textbf{Outperform Alternative Prediction Head.}
Besides the commonly used FC layer, we also evaluate DePT.
Although DePT can replace the FC layer and occasionally improve baseline performance, its design mainly targets the Base–New Tradeoff problem rather than robustness under missing modalities, leading to less consistent gains.
In contrast, our DPL provides more stable and adaptive improvements across different missing-modality patterns.
For example, as shown in \cref{table:dept-results}, when 50\% of text or images are missing, applying DePT to MAP reduces performance from 85.13\% $\rightarrow$ 84.02\% ($\Delta\%$=-1.11), while DPL improves the same baseline from 85.13\% $\rightarrow$ 87.08\% ($\Delta\%$=+1.85).
Under a severe missing rate (\eg, 90\% mixed-modality missing), DePT slightly enhances the FC baseline from 79.54\% $\rightarrow$ 79.97\% ($\Delta\%$=+0.43), whereas DPL further increases it to 81.98\% ($\Delta\%$=+2.44).
These results demonstrate that while DePT can serve as an alternative to the FC head, DPL achieves more reliable and consistent improvements across challenging missing-modality scenarios.

\subsection{Generalization and Robustness Analysis}

\noindent \textbf{Different Missing Rates.}
We conducted experiments to evaluate the robustness of the proposed DPL 
when trained on a dataset with predefined missing rate
and tested under varying missing rates.
Following previous prompt-based methods \cite{lee2023cvpr, hu2024deep}, 
we used the MM-IMDb dataset as an example, 
training with a 70\% missing rate and 
evaluating its performance across different testing missing rates. 
As shown in \cref{fig:fine-grained-missing}, 
when combined with DCP, 
our method DPL (red) consistently outperforms other altenatives, 
\eg, FC (blue) and DePT (yellow), 
across different missing cases and different testing missing rates.

\begin{figure*}[t]
    \label{fig:various-simu}
    \centering
    \includegraphics[width=0.9\linewidth]{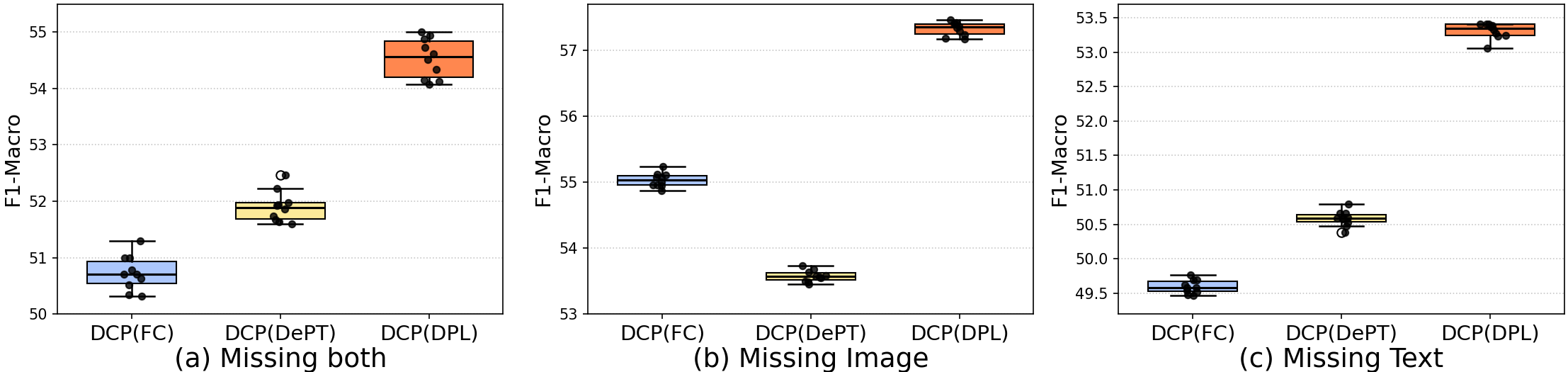}
    \caption{F1-Macro comparison of prediction heads (FC, DePT, DPL) integrated with the missing-aware DCP framework under three missing-modality settings: (a) both missing, (b) image missing, and (c) text missing. The proposed DPL achieves the highest F1-Macro across all cases, demonstrating superior robustness.}
\end{figure*}

\noindent \textbf{Varied Simulations.}
We also introduced a novel robustness evaluation, an aspect not previously explored in related works.
In missing modality simulation,
different random seeds can result in varying missing modalities pattens (\eg, missing image, missing text) 
and different combinations of missing modalities. 
As a result, the missing cases in the test set vary across simulations.
However, previous studies typically rely on a single simulation seed, 
limiting the assessment of model stability across different missing cases.
To comprehensively evaluate the robustness of our method, 
we calculated test performance across 10 different random simulations. 
The results, visualized in box plots in \cref{fig:various-simu}, 
demonstrate the effectiveness of our approach.
For both image-missing and text-missing case, 
our method (DPL) exhibits high consistency, as indicated by the narrow red boxes.
In the both-missing scenario, 
although our method shows a wider inter-quartile range (IQR), 
it still outperforms DePT (which exhibits outliers) and FC (which performs significantly worse than DPL). 
Furthermore, DPL achieves a higher upper-bound, 
improving the best-case performance of the base method, DCP.

\begin{table}[t]
    \centering
    \caption{F1-Macro comparison of different losses and prototype designs on Hateful-Memes under varied missing rates. The decomposed prototypes optimized by $L_{\text{DPL}}$ consistently achieves the best performance. The best results are highlighted in \textbf{bold}.}
    \footnotesize
    \adjustbox{max width=0.9\linewidth}{
    \begin{tabular}{c|cc|ccc}
        \hline 
           & & & \multicolumn{3}{c}{ w/o Prompt Tuning} \\
           \multirow{2}{*}{\makecell[c]{  Missing \\ rate $\eta$}} & \multicolumn{2}{c|}{ Train/Test} & $\mathcal{L}_{\text{DPL}}$ & $\mathcal{L}_{\text{ArcFace}}$ & $\mathcal{L}_{\text{DPL}}$  \\
           & Image & Text & un-decomp. & decomp. &  decomp. \\
          \hline 
           \multirow{3}{*}{$50 \%$}                         & $100 \%$ & $50 \%$  & 69.62 & 71.47 & \textbf{71.61}  \\
                                                            & $50 \%$ & $100 \%$  & 65.26 & 66.44 & \textbf{67.46}  \\
                                                            & $75 \%$ & $75 \%$   & 66.82 & 68.59 & \textbf{69.31}  \\
          \cline{1-6}
                                   \multirow{3}{*}{$70 \%$} &  $100 \%$ & $30 \%$ & 68.69 & 70.31 & \textbf{70.57} \\
                                                            & $30 \%$ & $100 \%$  & 64.27 & 65.16 & \textbf{65.62} \\
                                                            & $65 \%$ & $65 \%$   & 66.23 & 66.32 & \textbf{67.68} \\
          \cline{1-6}
                                   \multirow{3}{*}{$90 \%$} & $100 \%$ & $10 \%$  & 69.29 & 69.55 & \textbf{70.76}  \\
                                                            & $10 \%$ & $100 \%$  & 63.36 & 63.32 & \textbf{63.60} \\
                                                            & $55 \%$ & $55 \%$   & 66.36 & 64.22 & \textbf{66.40} \\
         \hline 
    \end{tabular}
    }
    \label{table:abl-decompose-loss}
\end{table}

\subsection{Ablation Study}

\noindent \textbf{Prototype Decomposition.}
We also explored another variation of the missing-case-aware prototypes, 
which are not decomposed into modality-specific ones.
To assess its impact, we conducted experiments on
MM-IMDb using the DCP with different ouput heads, 
and report the results in \cref{table:abl-decompose-loss}.
As shown in \cref{table:abl-decompose-loss}, 
the decomposed prototype (denoted as decomp.) achieves better performance 
compared with un-decomposed prototypes (denoted as un-decomp.),
which demonstrates that explicitly modeling modality-wise prototypes 
is crucial for improving model performance in missing modality scenarios.

\noindent \textbf{Loss term.} We conducted a set of ablation experiments to verify the effectiveness of the loss term $\mathcal{L}_{\text{PRC}}$.
The testing results on MM-IMDb for DCP with different output heads,
including our approach, are reported in \cref{table:abl-decompose-loss}, 
where $\mathcal{L}_{\text{ArcFace}}$ is presented in \cref{sec:loss}.
Tab. 2 shows that $\mathcal{L}_{DPL}$ consistently outperforms $\mathcal{L}_{\text{ArcFace}}$
across different missing cases.
Notably, the performance improvement of $\mathcal{L}_{\text{ArcFace}}$ is most pronounced.
in the both-missing scenario, 
compared to cases where only single modality is missing.
We attribute this to the fact that
both-missing scenario involves more decomposed prototypes in the $\mathcal{L}_{\text{PRC}}$, 
thus incorporating more views' information.

\begin{table}[t]
    \centering
    \caption{Comparison of MAP and the proposed DPL with and without the missing-aware (MA) mechanism, which explicitly selects prompts or prototypes based on the missing pattern. When the MA mechanism is disabled and replaced by minimum-entropy selection, DPL remains more robust, showing smaller performance degradation.}
    \adjustbox{max width=0.48\textwidth}{
    \begin{tabular}{cc|ccc|ccc}
        \hline 
        \multicolumn{2}{c|}{ } & \multicolumn{3}{c|}{MAP (CVPR'23)} & \multicolumn{3}{c}{Ours (proposed)} \\
        \multicolumn{2}{c|}{ Train/Test} & \multicolumn{3}{c|}{w/ Prompt Tuning} & \multicolumn{3}{c}{w/o Prompt Tuning} \\
        \multicolumn{2}{c|}{ $\eta$=\textcolor[HTML]{007acc}{50}/\textcolor[HTML]{0a8842}{70}/\textcolor[HTML]{B66F1B}{90}\% } & w/ FC & w/ FC & $\Delta$ & w/ DPL & w/ DPL & $\Delta$ \\
        Image &  Text  & w/o MA & w/ MA &   & w/o MA & w/ MA &  \\
        \hline 
        \textcolor[HTML]{007acc}{100}\% & \textcolor[HTML]{007acc}{50}\%  & 66.94 & 66.72 & -0.22 & 70.59 & 70.84  & +0.25 \\
        \textcolor[HTML]{007acc}{50}\%  & \textcolor[HTML]{007acc}{100}\% & 62.15 & 63.89 & +1.74 & 65.56 & 67.46  & +1.9 \\
        \textcolor[HTML]{007acc}{75}\% & \textcolor[HTML]{007acc}{75}\%   & 66.02 & 67.16 & +1.14 & 66.97 & 69.31  & +2.34 \\
        \cline{1-8}
        \textcolor[HTML]{0a8842}{100}\% & \textcolor[HTML]{0a8842}{30}\% & 68.77 & 68.76 & -0.01 & 70.60 & 70.57 & -0.03 \\
        \textcolor[HTML]{0a8842}{30}\% & \textcolor[HTML]{0a8842}{100}\% & 61.88 & 60.87 & -1.01 & 64.80 & 65.62 & +0.82  \\
        \textcolor[HTML]{0a8842}{65}\% & \textcolor[HTML]{0a8842}{65}\%  & 62.98 & 64.99 & +2.01 & 66.47 & 67.68 & +1.21  \\
        \cline{1-8}
        \textcolor[HTML]{B66F1B}{100}\% & \textcolor[HTML]{B66F1B}{10}\% & 67.40 & 67.57 & +0.17 & 70.63 & 70.76 & +0.13 \\
        \textcolor[HTML]{B66F1B}{10}\% & \textcolor[HTML]{B66F1B}{100}\% & 57.41 & 62.88 & +5.47 & 63.83 & 63.60 & -0.23 \\
        \textcolor[HTML]{B66F1B}{55}\% & \textcolor[HTML]{B66F1B}{55}\%  & 59.55 & 62.07 & +2.52 & 66.33 & 66.40 & +0.07 \\
        \hline
         
    \end{tabular}
    }
    \label{table:nonfilter-results}
\end{table}

\noindent \textbf{Effect of Missing-Aware Mechanism.} The default DPL adopts a missing-aware mechanism similar to MAP, which selects corresponding prototypes for logits computation based on detected missing modalities.
To test its robustness, we further disable this mechanism and compare MAP (missing-aware prompts) with DPL (missing-aware prototypes).
When missing-awareness is disabled, we select the most confident logits using minimum entropy among candidates, as shown in \cref{table:nonfilter-results}.
DPL consistently benefits from the missing-aware design (positive $\Delta$) and shows smaller performance drops than MAP when the mechanism is removed, indicating stronger inherent robustness.

\begin{figure}[ht]
    \centering
        \includegraphics[width=0.8\linewidth]{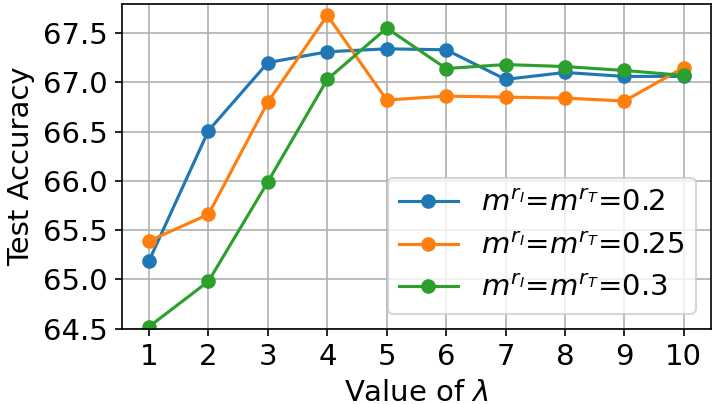}
    \caption{Test accuracy under a 70\% missing rate in a mixed-missing scenario (\eg, either image or text could be missing), showing the influence of $\lambda$, $m^{r_{I}}$ and $m^{r_{T}}$ with fixed $m^{c}$=0.15.
    }
    \label{fig:coef-line}
\end{figure}

\noindent \textbf{Hyperparameters Analysis.} As shown in \cref{fig:coef-line}, we examine the influence of key hyperparameters on test accuracy under a mixed-missing scenario.
Across different angular margin settings ($m^{r_I}$ and $m^{r_T}$) with a fixed $m^{c}$ = 0.15, the overall performance trends remain consistent.
A small $\lambda$ results in clear degradation, indicating insufficient regularization for balancing the class-level and relation-level objectives.
As $\lambda$ increases, the performance steadily improves and reaches a stable plateau, demonstrating the robustness of our framework to a broad range of $\lambda$ values.
\section{Conclusion}
In this work, we addressed the critical challenge of improving the robustness of Vision–Language Transformers under missing modalities. We proposed a decoupled prototype-based prediction head that functions as a strong standalone classifier while also providing complementary gains when integrated with existing prompt-based methods. To support effective prototype learning, we decompose missing-aware prototypes into modality-specific components and introduce two complementary losses, namely the ArcFace margin loss and the Prototype Relational Contrastive loss, to regularize inter-class and intra-class prototype relations. Extensive experiments across diverse missing-modality scenarios and datasets verify that our approach consistently enhances robustness and remains fully compatible with existing prompt-based framework.
{
    \small
    \bibliographystyle{ieeenat_fullname}
    \bibliography{main}
}


\end{document}